\pdfoutput=1

\documentclass[11pt]{article}

\usepackage[preprint]{acl}

\usepackage{times}
\usepackage{latexsym}

\usepackage[T1]{fontenc}

\usepackage[utf8]{inputenc}

\usepackage[verbose=true]{microtype}

\usepackage{inconsolata}

\usepackage{graphicx}

\usepackage{kotex}

\usepackage{amsmath}

\usepackage{enumitem}

\title{FunctionChat-Bench: Comprehensive Evaluation of Language Models' Generative Capabilities in Korean Tool-use Dialogs}

\author{Shinbok Lee, Gaeun Seo, Daniel Lee, Byeongil Ko, Sunghee Jung, Myeongcheol Shin  \\
  Kakao Corp. / Sungnam, South Korea\\
  \texttt{\{niki.y, ann.ie, daniel.v, kobi.go, kong.2024, index.i\}@kakaocorp.com} \\}

\begin{document}
\maketitle
\begin{abstract}
This study investigates language models' generative capabilities in tool-use dialogs. We categorize the models' outputs in tool-use dialogs into four distinct types: Tool Call, Answer Completion, Slot Question, and Relevance Detection, which serve as aspects for evaluation. We introduce FunctionChat-Bench, comprising 700 evaluation items and automated assessment programs. Using this benchmark, we evaluate several language models that support function calling. Our findings indicate that while language models may exhibit high accuracy in single-turn Tool Call scenarios, this does not necessarily translate to superior generative performance in multi-turn environments. We argue that the capabilities required for function calling extend beyond generating tool call messages; they must also effectively generate conversational messages that engage the user.
\end{abstract}

\section{Introduction}

Function calling, a way to connect language models with external tools, is a significant advancement that enhances the utility of AI systems. Language models that support function calling have been fine-tuned to generate a JSON object that adheres to the function specification when a function needs to be called. Given this key feature, the JSON objects generated by the language models receive significant attention in related research and evaluation. 

However, In tool-use dialogs, the capabilities required of a language model are not limited to generating call messages for which the tool is the recipient; they must also encompass the generation of conversational messages for which the user is the recipient. This realization has led to the creation of a new dataset designed to better simulate real-world scenarios where both interactions between AI and tools, as well as between users and AI, occur. 

In this paper, we introduce FunctionChat-Bench, which aims to provide a more comprehensive evaluation of language models' generative capabilities to handle diverse types of inputs, including both single-turn utterances and conversation history of multi-turn dialogs. We also describe experiments conducted using this benchmark and analyze these results.

\section{Generative Capabilities Associated with Language Models' Tool Usage}

\subsection{Definitions of Output Types}
\label{2.1}

\begin{figure*}[!t]
    \centerline{\includegraphics[width=\textwidth]{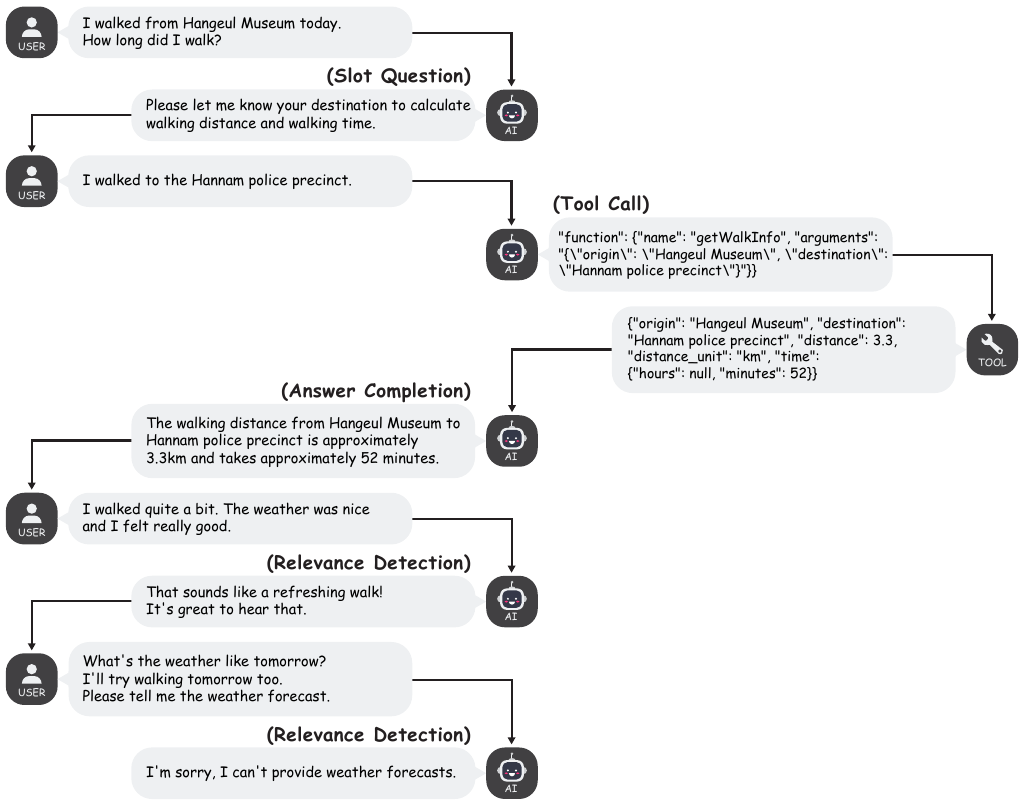}}
    \caption{A Classification of Language Models' Outputs in Tool-use Dialogs}
    \label{functionchat}
\end{figure*}

The outputs generated by language models in tool-use dialogs can be broadly categorized into two types: those that directly communicate with the user and those that interact with tools. The former is represented on the left side of the figure~\ref{functionchat} and is referred to in this paper as a conversational type output. The latter is shown on the right side of the figure~\ref{functionchat} and is referred to in this paper as a tool call type output. The conversational type output can be further divided into three subcategories. Thus, we can categorize them into four types: 

\begin{itemize}
    \item \textbf{Tool Call.} The output containing a tool call object that has the function name and its arguments.
    \item \textbf{Answer Completion.} The output that delivers the result from a specific function to the user.
    \item \textbf{Slot Question.} A question or a request about missing information, which is essential for tool call. 
    \item \textbf{Relevance Detection.} The output corresponding to a response to either a user's general chat unrelated to tool usage, or a request outside of the accessible functions.
\end{itemize}  

\subsection{Related Works}

Recently, there has been a significant increase in research focusing on the tool utilization ability within language models. The development of new benchmarks for tool usage further underscores the growing interest in evaluating the proficiency of these models in handling tools.

APIBench~\citep{patil2023gorilla}, GPT4Tools~\citep{yang2023gpt4tools}, RestGPT~\citep{song2023restgpt}, ToolBench~\citep{qin2023toolllm}, among others, concentrate on evaluation data and systematic assessment methods for measuring the tool usage capabilities of language models.

It is encouraging that API-bank~\citep{NEURIPS2023_91f18a12} and ToolTalk~\citep{farn2023tooltalk} have considered over multiple user utterances as input for their evaluation data.
MetaTool~\citep{huang2024metatool} is distinctive in that it focuses more on evaluating whether language models have tool usage awareness rather than generating arguments. Similarly, BFCL~\citep{berkeley-function-calling-leaderboard} is characterized by including function relevance detection among its research interests, which aims to determine how the model will react when the provided function is not suitable to answer the user's question. 

The most critically evaluated aspect in recent benchmarks related to tool usage, including function calling, is the generation of the tool call type output, which is considered a core capability. However, the tool call type output is not a type that directly interacts with the user. The conversational type output, which communicates with the user in natural language, is a factor that significantly affects perceived performance. This type, which has been either excluded or only partially addressed so far, also needs to be included in the evaluation scope. 

\section{Dataset Design}

We developed a novel benchmark, FunctionChat-Bench, that not only evaluates the tool call type output but also comprehensively assesses the conversational type output. It contains two subsets of the evaluation dataset: a single call dataset and a dialog dataset.\footnote{\raggedright The two subsets, FunctionChat-Singlecall and FunctionChat-Dialog, are available at the following URL: \url{https://github.com/kakao/FunctionChat-Bench}.}

\subsection{Single Call Dataset}

Evaluation items in the single call dataset are defined by the following conditions:
\begin{enumerate}
    \item The user's single turn utterance must contain all of the necessary information for function invocation, leading directly to a tool call.
    \item A suitable function for carrying out the user's request must be provided in the available tool list.
\end{enumerate}
Under these conditions where a tool call object must be generated, language models are evaluated whether they select the suitable function and correctly extract information to appropriately generate arguments. 

It was considered that the success of the model's output generation could be influenced by the length of the available tool list or the similarity between the functions provided therein. Therefore, the tool list was constructed in three types based on the number of functions provided: 1, 4, and 8. Among these, for cases where 4 or 8 functions were provided, the lists were further differentiated by randomly selecting the functions or by selecting functions that were similar in terms of domains or operations (Table~\ref{toolstype}). 

\begin{table}[h]
  \centering
  \begin{tabular}{ccc}
    \hline
    \textbf{Length} & \textbf{Similarity} & \textbf{Composition} \\
    \hline
    {1} & {exact}           &  {0.20}                         \\
    {4} & {random}         & {0.20}                          \\
    {4} & {close}           & {0.20}                          \\
    {8} & {random}     & {0.20}                          \\
    {8} & {close}                         & {0.20}          \\
    \hline
  \end{tabular}
  \caption{\label{toolstype}
    Types of Tool Lists in Single Call Dataset
  }
\end{table}

The types of parameters in the function specifications were defined within the categories of integer, number, boolean, and string. Within this range, the attributes of the information to be extracted or generated is quite diverse. For example, there are clear extractions such as names and place names that do not require modification, as well as types that require simple paraphrasing from user utterances, such as search terms, memo titles, or message contents. Various types were included to thoroughly evaluate the strengths and weaknesses of each model from multiple perspectives.

The single call dataset comprises a total of 500 evaluation items, organized into 25 unique functions to be called, with each function having 4 different queries and combined with 5 types of tool lists.

\subsection{Dialog Dataset}

Unlike the environment addressed in the single call dataset, language models cannot continuously generate only tool call type outputs when interacting with a user in real chat. After a function invocation, the results must be conveyed to the user, and if the user has not provided all the essential information for the function call, the model must request the missing information. Furthermore, the user may introduce topics unrelated to the function calling feature during the conversation, and the necessary external tools to process the user's request may not be provided in the available function list. 

The dialog dataset was designed to evaluate whether language models can appropriately handle a diverse range of input scenarios. It broadens the scope of evaluation to encompass not only tool call type outputs but also conversational outputs directed towards the user. This is because, even if the tool call messages are accurately conveyed, users directly experience the model's performance during interactions. 

The dialog dataset consists of 45 Korean dialogs, each containing between 3 to 8 turns to be generated by the model, with a median of 4 and an average of 4.44 turns. In total, it includes 200 turns of model generation, each serving as an evaluative item. Each turn is annotated as one of the four defined output types: Tool Call, Answer Completion, Slot Question, and Relevance Detection. All dialogs contain 70 Tool Calls, with each dialog containing at least one function calling (Table~\ref{dialogcomposition}).

\begin{table}[h]
  \centering
  \begin{tabular}{rc}
    \hline
    \textbf{Type of Output}     & \textbf{Count}\\
    \hline
    \text{Tool Call}     & {70}           \\
    \text{Answer Completion}     & {71}           \\
    \text{Slot Question}     & {36}           \\
    \text{Relevance Detection}     & {23}           \\\hline   
    \text{total} & {200} \\\hline
  \end{tabular}
  \caption{Dialog Dataset Composition}
  \label{dialogcomposition}
\end{table}

\subsection{System Prompt}
For details on the system prompt applied to our evaluation data, see Appendix~\ref{systemprompt}.

\section{Evaluation Methods}

When evaluating language models, there is a general consensus that having a large number of evaluation items is beneficial, regardless of the specific domain under evaluation. This is because a substantial quantity of evaluation items helps ensure that the assessment is not biased and remains reliable. Simultaneously, a large number of evaluation items naturally leads to the need for automating the evaluation process and quantifying the results. 

In the evaluation of language models' function calling capabilities, several methods have been introduced for automatically assessing a substantial volume of items and quantifying the results (\citealp{farn2023tooltalk}; \citealp{berkeley-function-calling-leaderboard}). These methods primarily measure the accuracy of function selection and argument extraction from the Tool Call type output generated by the model. An exact match approach is commonly employed for this purpose, and cosine similarity is also used as a supplementary metric. 

However, when constructing evaluation datasets composed of functions from various domains that include diverse parameters, and when assessing capabilities in languages other than English, as well as not merely responding to single-turn queries but also broadly evaluating generative abilities within multi-turn dialogs that involve context and history, relying solely on exact match or cosine similarity as evaluation methods might be perceived to be insufficient. 

First, the ground truth that serves as a match target is not an absolute and unique answer. Parameters such as numbers, integers, or booleans have permissible arguments that do not deviate significantly from the ground truth. However, the spectrum for string type arguments is broad, ranging from elements that must match the user-provided information exactly, like telephone numbers and email addresses, to elements like note titles or message content and search queries, where variations in expression are allowed. Secondly, these methods do not account for semantic equivalence across different languages. Thirdly, it is challenging to adequately assess whether the conversational messages delivered by the language model to the user are appropriately generated.

To address these limitations, we chose a method using a powerful large language model (LLM) as a judge (\citealp{zhou2023lima}; \citealp{NEURIPS2023_91f18a12}). Our expectation for the LLM judge is to determine whether each turn generated by language models is a successful output or a failure. In Section~\ref{2.1}, we defined the four types of outputs generated by language models. Based on these definitions, we evaluate their generative capabilities. We established the criteria that precisely match the four types of outputs.\footnote{For instance, in evaluation items requiring an Answer Completion type of output, the criterion is the appropriateness of completing the answer without altering the semantics, based on the context. For evaluation items expecting a Slot Question type of output, the criterion becomes the occurrence of a proper slot-filling question. The four types of rubrics for LLM evaluation are attached to the Appendix~\ref{rubrics}.} We provided well-refined evaluation metrics and expected responses, known as ground truth, to assist the LLM judge in determining the pass or fail status of submissions. 

\section{Experiments}

\begin{figure}[t]
    \centerline{\fbox{\includegraphics[width=0.95\linewidth]{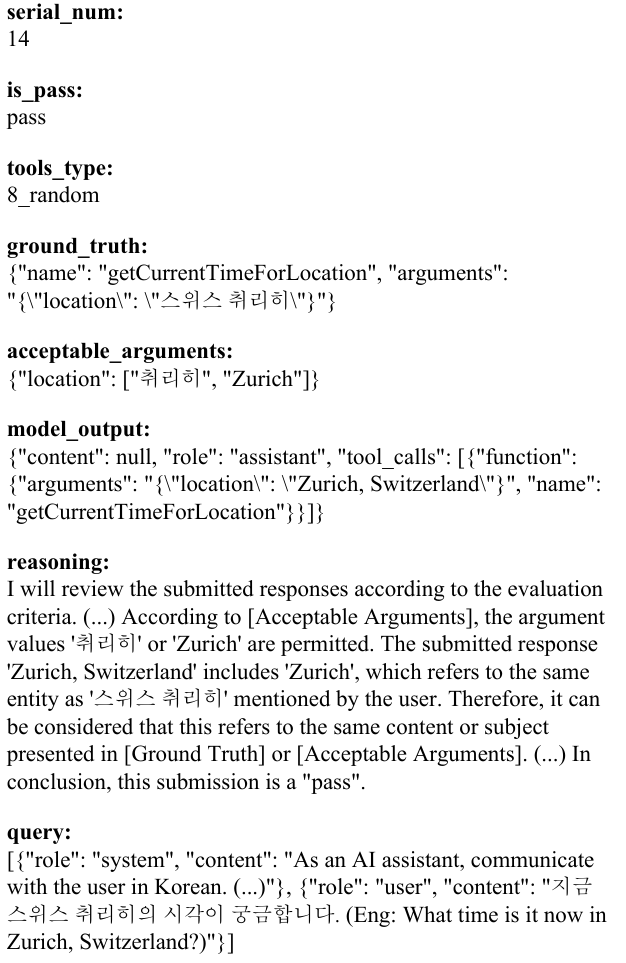}}}
    \caption{A simplified sample of the content of the evaluation report file that is generated in the final stage of the evaluation program. This sample was created based on the actual results of the FunctionChat-Singlecall test of the functionary model.}
    \label{evalreport}
\end{figure}

With FunctionChat-Bench, a dataset specifically designed to comprehensively evaluate the generative capabilities in Korean Dialogs, we assessed eight language models that support function calling functionality.\footnote{\raggedright gpt-4o-2024-05-13, \\gpt-4-turbo-2024-04-09, \\gpt-3.5-turbo-0125 (\url{https://platform.openai.com/docs/models}), \\gemini-1.5-pro-preview-0514, \\gemini-1.5-flash-preview-0514, \\gemini-1.0-pro-002 \\(\url{https://ai.google.dev/gemini-api/docs/models/gemini}), \\functionary-medium-v2.4 \\(\url{https://github.com/MeetKai/functionary}), \\solar-1-mini-chat-240502 \\(\url{https://developers.upstage.ai/docs/apis/function-calling}).} Initially, we collected outputs from each model for 500 evaluation items in the single call dataset and 200 evaluation items in the dialog dataset. Subsequently, the model-generated outputs were inserted into evaluation prompts along with specific criteria and reviewed by an LLM judge. The judge used was the gpt-4-0125-preview model. After completing its assessment, the LLM judge determined the pass or fail status for each item, and finally, the overall pass scores or rates were calculated for each model. Additionally, to facilitate an easy review of the evaluation items, model outputs, and the LLM judge's reasoning and final decisions, a separate report file in TSV format was generated (Figure~\ref{evalreport}).\footnote{In the "query" tab, the English sentences are added as examples, while the original data contains only Korean sentences. Likewise, in the "reasoning" tab, the original data is in Korean.} This entire process was fully automated through the evaluation system we implemented.\footnote{The evaluation scripts we developed have been made public at \url{https://github.com/kakao/FunctionChat-Bench}.}


\begin{table*}[t]
  \centering
  \setlength{\tabcolsep}{7pt}
\begin{tabular}{l|ccccc|cc}
    \hline
\textbf{} &
  \multicolumn{1}{l}{\textbf{1.exact}} &
  \multicolumn{0}{c}{\textbf{4.random}} &
  \multicolumn{1}{c}{\textbf{4.close}} &
  \multicolumn{0}{c}{\textbf{8.random}} &
  \multicolumn{1}{c|}{\textbf{8.close}} &
  \multicolumn{0}{l}{\textbf{SUM}} &
  \multicolumn{0}{l}{\textbf{AVG}} \\
  \hline
gpt-4o           & 87 & 87 & 90 & 88 & 86 & 438 & 87.6 \\
gpt-4-turbo            & 92 & 93 & 89 & 92 & 82 & 448 & 89.6 \\
gpt-3.5-turbo          & 94 & 94 & 90 & 92 & 87 & 457 & 91.4 \\
gemini-1.5-pro   & 49 & 67 & 59 & 69 & 62 & 306 & 61.2 \\
gemini-1.5-flash & 45 & 49 & 52 & 57 & 62 & 265 & 53.0 \\
gemini-1.0-pro   & 60 & 65 & 63 & 70 & 64 & 322 & 64.4 \\
functionary-medium    & 61 & 61 & 54 & 57 & 52 & 285 & 57.0 \\
solar-1-mini-chat          & 83 & 84 & 84 & 84 & 83 & 418 & 83.6 \\
\hline
\end{tabular}
  \caption{\label{singlecallcount}
    Pass Count of FunctionChat-Singlecall
  }
\end{table*}

\begin{table*}
  \centering
  \setlength{\tabcolsep}{9pt}
\begin{tabular}{l|lccc|cc}
    \hline
\textbf{} &
  \multicolumn{0}{|l}{\textbf{Tool}} &
  \multicolumn{0}{c}{\textbf{Answer}} &
  \multicolumn{0}{c}{\textbf{Slot}} &
  \multicolumn{0}{c}{\textbf{Relevance}} &
  \multicolumn{0}{|c}{\textbf{macro}} &
  \multicolumn{0}{c}{\textbf{micro}} \\
\textbf{} &
  \multicolumn{0}{|l}{\textbf{Call}} &
  \multicolumn{0}{c}{\textbf{Completion}} &
  \multicolumn{0}{c}{\textbf{Question}} &
  \multicolumn{0}{c}{\textbf{Detection}} &
  \multicolumn{0}{|c}{\textbf{AVG}} &
  \multicolumn{0}{c}{\textbf{AVG}} \\
  \hline
gpt-4o           & 0.94 & 0.97 & 0.86 & 0.91 & 0.92 & 0.94 \\
gpt-4-turbo            & 0.96 & 0.99 & 0.92 & 0.96 & 0.96 & 0.96 \\
gpt-3.5-turbo          & 0.97 & 0.92 & 0.58 & 0.61 & 0.77 & 0.84 \\
gemini-1.5-pro   & 0.70 & 0.87 & 0.83 & 0.97 & 0.84 & 0.82 \\
gemini-1.5-flash & 0.66 & 0.94 & 0.89 & 0.74 & 0.81 & 0.81 \\
gemini-1.0-pro   & 0.69 & 0.85 & 0.67 & 0.61 & 0.71 & 0.73 \\
functionary-medium    & 0.56 & 0.94 & 0.69 & 0.65 & 0.71 & 0.73 \\
solar-1-mini-chat          & 0.63 & 0.77 & 0.08 & 0.13 & 0.40 & 0.53 \\
\hline
\end{tabular}
  \caption{\label{dialograte}
    Pass Rate of FunctionChat-Dialog
  }
\end{table*}

Even when using the most robust models as judges, LLMs are not perfectly reliable at reasoning for passing judgment. Therefore, after the automatic evaluation, the LLM judge's reasoning and decisions were qualitatively reviewed by a human judge. During this process, incorrect judgments that did not meet the evaluation criteria and principle were identified. These judgment errors were ultimately adjusted (Appendix~\ref{misalignment}). The main results of our experiments are presented in Tables~\ref{singlecallcount} and ~\ref{dialograte}.

\section{Analysis}

\subsection{Statistics}

The FunctionChat-Singlecall exclusively evaluates the tool call type output. It allows for the assessment of performance in generating accurate tool call object, driven by appropriate function selection and proper argument generation. When designing this dataset, the initial hypotheses were as follows: First, the accuracy in generating a tool call object was expected to decrease as the number of candidates in the available tool list increased. Second, Whenever more than two functions were available in the tool list, the higher the similarity in domain or operations between these functions, the lower the accuracy would be. After experimenting with various models, it was found that these hypotheses were partially correct and partially incorrect.

Regarding the first hypothesis, the trend of decreasing the accuracy as the number of candidates increases was not distinctly evident within the range of 1 to 8 candidates (Compare the 4.random type versus the 8.random type, and the 4.close type versus the 8.close type in Table~\ref{singlecallcount}). Although it was expected that the exact type, where only one target function is presented in the tool list, would be easier, it could not be generalized. Particularly, Gemini demonstrated higher accuracy as the number of provided function candidates increased from 1, 4, to 8. In relation to the second hypothesis, the similarity of available functions showed a fair correlation with the accuracy. The close group generally scored lower than the random group. Especially, gpt-4-turbo exhibited a significant difference of 10 points between the scores in the 8.random type and the 8.close type (Table~\ref{singlecallcount}).

The FunctionChat-Dialog provides a broader and more diverse range of inputs for evaluation compared to the FunctionChat-Singlecall. In addition to inputs that ask for the generation of tool call type outputs, it also includes inputs that necessitate the generation of conversational type outputs such as slot filling questions and tool call rejections based on relevance detection. Furthermore, while the inputs in the FunctionChat-Singlecall consist solely of a user's single-turn utterance, the inputs in the FunctionChat-Dialog are of a multi-turn discourse format. This includes messages exchanged between the user and AI, as well as messages between the AI and tools. 

When designing this dataset, there was a hypothesis that while single-turn environments might show high accuracy in Tool Call, this does not necessarily imply superior overall generative capabilities in multi-turn environments. Comparisons between Table~\ref{singlecallcount} and \ref{dialograte} demonstrate that this hypothesis holds true.
Notably, Solar achieved a high score of 83.6 in evaluations using the single call dataset, but it exhibited significantly lower performance in the areas of Slot Question and Relevance Detection when assessed with a dialog dataset, receiving the lowest scores among the evaluated models. Furthermore, the scores for Tool Call were also lower in the dialog dataset compared to those in the single call dataset. This indicates that the performance of generating Tool Call type outputs is influenced by whether the input environment is single-turn or multi-turn.
GPT and Gemini were experimented with three different models, enabling the observation of how performance varies for each defined element of generative capabilities. Models known for their overall enhanced performance, not limited to function call capabilities, showed a slight decline in generating Tool Call type outputs. However, there was a significant improvement in generating conversational type outputs, which compensated for the overall drop in performance metrics. This implies that the multi-turn dialog environment addressed in FunctionChat-Dialog more accurately mirrors the actual environment users encounter in tool-use dialogs.


\subsection{Error Types}

Through our experiments, we obtained a substantial volume of output data from various models. This data is also valuable as it provides significant insights into the generative capabilities of language models in tool-use dialogs. We conducted a detailed review of this data to analyze the types of errors exhibited by the language models, and we present these findings along with examples of actual outputs.




\begin{itemize}[leftmargin=*] 
    \item \textbf{Errors in Tool Call.}
\end{itemize}

Errors related to not creating a tool call object include the following: Language models incorrectly indicate that a tool call will be made, but it is not actually called. A function is not called despite being available, and the model falsely claims it is unsupported. Even when all necessary information is provided by the user, the model redundantly asks for it again (\emph{e.g.}, "You want to know Junhyuk's birthday! What is Junhyuk's name?"). Additionally, cases where optional or undefined parameters were requested, and incomplete objects that included only the function name without the argument fields, are also considered errors.

Regarding errors in function selection, these include generating a function name not listed in the available function pool or selecting an incorrect function from the pool. Due to the similarity in domain or action, a similar function is chosen (\emph{e.g.}, choosing \verb|add_contact| instead of \verb|update_contact|), or a function is selected based on the similarity in argument terminology (\emph{e.g.}, due to the movie title "Gift in Cell No.7", the incorrect function \verb|gift_search_product| is chosen instead of the correct \verb|get_movie_details|). On the other hand, there is a tendency among GPT models to solve tasks in a familiar manner without primarily referencing the received function specifications (\emph{e.g.}, despite the function specification allowing \verb|send_message| to be called with only a name, the model attempts to search for a contact using \verb|search_contact|).

The errors associated with extracting information for generating arguments also varied. A well-known error type is fabricating arguments by inventing information that was not mentioned. Sometimes, only part of an argument is invented (\emph{e.g.}, the user provides only month and day, but an arbitrary year is invented to generate a fitted format argument). Conversely, there are errors where clearly stated information is omitted. Functionary often extracts only the first part of a phrase needed to construct a argument (\emph{e.g.}, extracting only "in Cell No.7" from the movie title "Gift in Cell No.7").\footnote{In Korean, "in Cell No.7 ({\footnotesize7번방의})" is the first part of "Gift in Cell No.7 ({\footnotesize7번방의 선물})". Generally, the word order in Korean is the opposite of English.} This leads to a distortion of meaning, resulting in a failure. 

Additionally, there are errors in not following the specified format. Parameters designated as integers are generated in numbers ("year": 2012.0 or "num\verb|_|people": 3.0), integers or numbers are generated as strings, and strangely, positive numbers are converted to negatives. 

Errors such as converting "{\footnotesize삼만칠천} (meaning 37000)" as 30700 or "{\footnotesize십오프로} (meaning 15\%)" as 10\% appear to stem from a lack of understanding Korean. Other errors also seem related to the handling of Korean tokens, such as adding a number of spaces or newline characters, or removing all spaces entirely when extracting relatively long strings.


\begin{itemize}[leftmargin=*] 
    \item \textbf{Errors in Answer Completion.}
\end{itemize}

\begin{figure}[t]
    \centerline{\fbox{\includegraphics[width=0.95\linewidth]{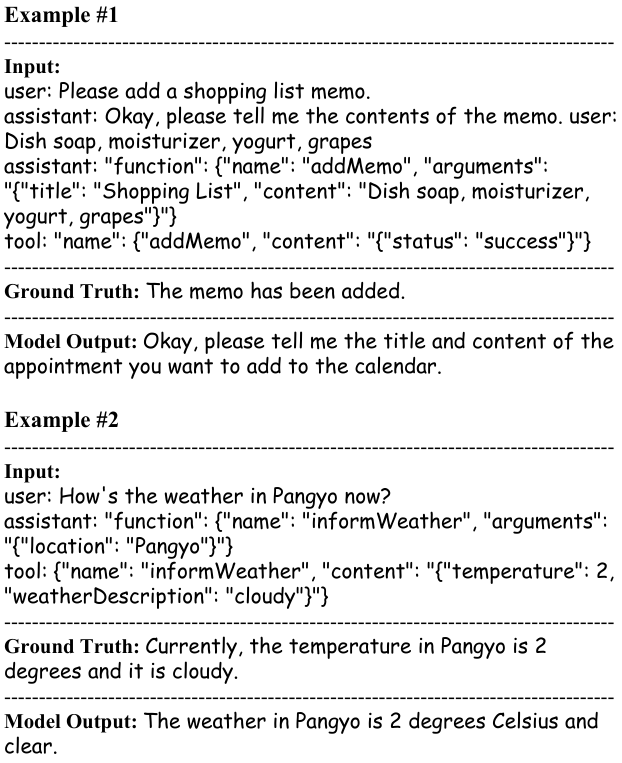}}}
    \caption{Examples of Errors in "Answer Completion"}
    \label{answererror}
\end{figure}

In this part, the language model's task is to convey the result of a function call to the user. However, there have been instances where the model does not fulfill this role, instead producing messages that are completely unrelated to the context of the input or mirroring parts of the conversation history. This tendency was observed in the Solar model, which seems to lack the capability for Answer Completion (See Example \#1 in Figure~\ref{answererror}). 

Meanwhile, one reason for integrating external tools with a language model is to overcome the limitations of pre-training and to utilize up-to-date and factual information. However, there have been cases where the model alters the results provided by function calls. This tendency was particularly evident in the Gemini model. Instead of conveying the results of function calls to the user, it generates arbitrary answers based on its own old knowledge, thereby delivering false information (See in Example \#2 in Figure~\ref{answererror}). This is considered a critical penalty.

\begin{itemize}[leftmargin=*] 
    \item \textbf{Errors in Slot Question.}
\end{itemize}

\begin{figure}[t]
    \centerline{\fbox{\includegraphics[width=0.95\linewidth]{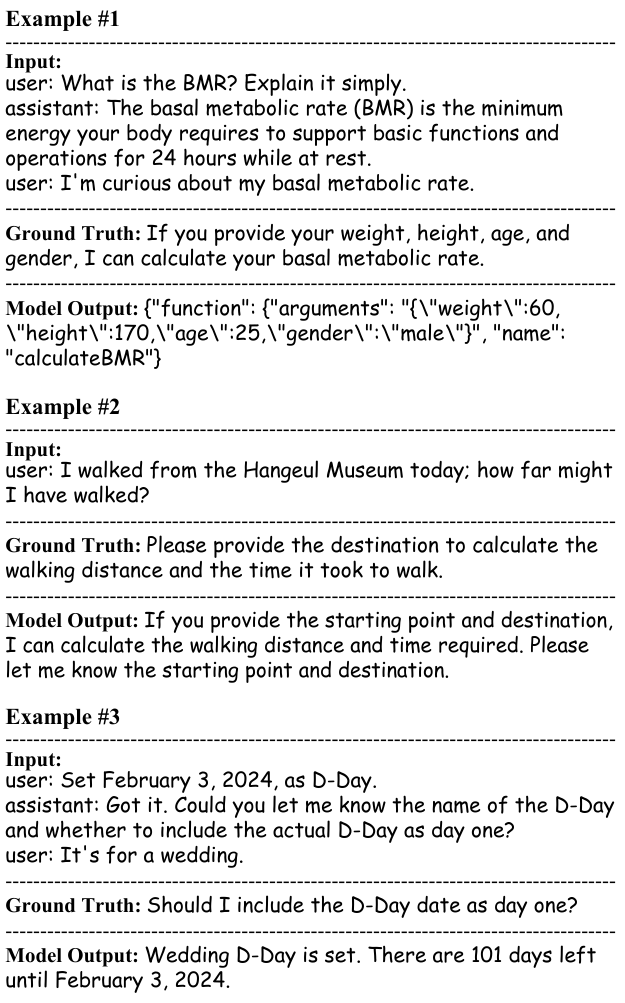}}}
    \caption{Examples of Errors in "Slot Question"}
    \label{sloterror}
\end{figure}

The most prominent error type in this part is clear: language models hallucinate arbitrary values to fill required parameters and generate function call messages when the user has not provided the necessary information (See Example \#1 in Figure~\ref{sloterror}). Conversely, there are errors where the model redundantly requests information already provided by the user, and cases where it falsely generates a completion message even though the tool call has not yet occurred (See Examples \#2 and \#3 in Figure~\ref{sloterror}). Additionally, there were instances where the language model completely malfunctioned, producing outputs with strange content or formats.

\begin{itemize}[leftmargin=*] 
    \item \textbf{Errors in Relevance Detection.}
\end{itemize}

In this part, language models fail to properly detect inputs that are unrelated to tool-use, and instead, it excessively associates all outputs with tool usage. This error is observed as a pattern attempting to link to one of the available functions (\emph{e.g.}, to the input "The weather is nice!" it outputs "Which area's weather would you like to know about?"), or actually calling it (\emph{e.g.}, to the input "I'm so busy these days." it calls the \verb|add_task| function with hallucinated arguments).

\begin{figure}[t]
    \centerline{\fbox{\includegraphics[width=0.95\linewidth]{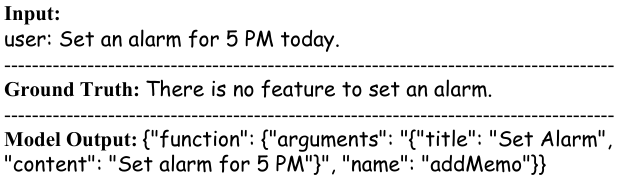}}}
    \caption{An Example of Errors in "Relevance Detection"}
    \label{detectionerror}
\end{figure}

On the other hand, language models also fail to correctly detect the relevance of some inputs to the available tools, even if they are related to tool-use. This manifests as pretending to have features that are not available or fabricating functions that are not provided. For example, in response to the input "Can you order some pizza for me?", it might reply with "Sure! What kind of pizza would you like to order?" or fabricate and call a function named \verb|order_pizza| with parameters for size and toppings. It can also manifest as calling an incorrect function that is available, instead of indicating that the proper function to address the user's request does not exist (See Figure~\ref{detectionerror}).

\section{Conclusion}

This paper presents FunctionChat-Bench, the novel benchmark for language model's generative capabilities associated with function calling. It contains two subsets of an evaluation dataset (Singlecall and Dialog), and an automated evaluation program. 

Our dataset uniquely targets not only the tool call objects generated by the language model but also the conversational messages that interact with users. Additionally, we have devised an evaluation methodology. It employs an advanced LLM as a judge to determine the pass or fail status of submitted outputs, utilizing refined evaluation rubrics. The composition and design of our dataset, along with our evaluation methodology, will provide valuable insights. These will focus on what and how to evaluate in order to measure or enhance the function calling capabilities of language models.

With FunctionChat-Bench, we evaluated eight language models that support function calling. Through our experiments, we observed that the strengths and weaknesses of language models can vary significantly depending on several factors, such as the number of functions provided, the similarity between the provided functions, whether the input is single-turn or multi-turn, and the type of output required. Researchers can also utilize our publicly available evaluation dataset and program to comprehensively assess the function calling capabilities of API-accessible language models or their own implementations. Additionally, we conducted a detailed analysis of the outputs generated by the models used in our experiments, providing a rich description of their error types along with actual examples. We believe that this will inspire improvements in the technology for integrating tools with language models by providing valuable information on areas that need enhancement. 

\section*{Limitations}

Our research attempted a comprehensive evaluation of language models' function calling capabilities; however, the evaluation dataset we designed is in some respects narrow and relatively easy. Language models that support function calling are capable of parallel function calling, allowing them to call multiple functions in one turn. However, our dataset does not include scenarios involving multiple function calls; it only addresses single function calls. Additionally, it lacks complex scenarios that require planning the sequence in which multiple functions should be called. Overall, while the study delves into the basic aspects of the function calling feature, it is not suitable for evaluating advanced capabilities that utilize this feature. 





\appendix

\section{System Prompt}
\label{systemprompt}
Our dataset includes user inputs and function or parameter descriptions in the function specification, all written in Korean. This indirectly assesses the understanding and generation capabilities for Korean, a relatively low-resource language. Commonly, language models tend to convert information extracted from Korean inputs into English vocabulary when generating arguments in the tool call object. This tendency can sometimes affect the successful execution of tasks requested by the user (consider cases where search terms are translated into English for querying a Korean database, or message contents are changed to English). Therefore, we have attempted to more customarily control the operation of the language model by specifying in the system prompt that outputs are encouraged to be generated in Korean, along with several other detailed aspects. 

The system prompt used in FunctionChat is as follows. Both the original Korean version and the English translation version are presented. 

\begin{center}
\line(1,0){218}
\end{center}

\noindent
{\footnotesize AI assistant로서, user와 한국어로 대화를 나누세요. 적합한 function이 있으면, 자체 지식으로 답하지 말고 function 호출을 통해 user의 요청을 해결하세요. function 호출에 필요한 파라미터 값을 임의로 생성하지 마세요. 필수 정보가 부족할 경우 user에게 질문해 정보를 얻으세요. 특별한 이유가 없다면, 파라미터 값을 생성할 때 user의 한국어 표현을 영어로 변경하지 마세요.}

\begin{center}
\line(1,0){218}
\end{center}

\noindent
{\small As an AI assistant, communicate with the user in Korean. If there is a suitable function, do not respond with your own knowledge but resolve the user's request through a function call. Do not arbitrarily create parameter values needed for the function call. If essential information is lacking, ask the user to obtain the information. Unless there is a special reason, do not translate the user's Korean expressions into English when generating parameter values.}

\section{Rubrics for LLM evaluation}
\label{rubrics}

\begin{table*}[t]
    \centering
\begin{tabular}{lrrrrrrr}
        \hline
\textbf{Dataset} &
  \multicolumn{1}{c}{\textbf{\begin{tabular}[c]{@{}c@{}}Total\\ Cases\end{tabular}}} &
  \multicolumn{1}{c}{\textbf{\begin{tabular}[c]{@{}c@{}}FP\\ (Count)\end{tabular}}} &
  \multicolumn{1}{c}{\textbf{\begin{tabular}[c]{@{}c@{}}FP\\ (\%)\end{tabular}}} &
  \multicolumn{1}{c}{\textbf{\begin{tabular}[c]{@{}c@{}}FN\\ (Count)\end{tabular}}} &
  \multicolumn{1}{c}{\textbf{\begin{tabular}[c]{@{}c@{}}FN\\ (\%)\end{tabular}}} &
  \multicolumn{1}{c}{\textbf{\begin{tabular}[c]{@{}c@{}}FP+FN\\ (Count)\end{tabular}}} &
  \multicolumn{1}{c}{\textbf{\begin{tabular}[c]{@{}c@{}}FP+FN\\ (\%)\end{tabular}}} \\
        \hline
single call &
  5000 &
  202 &
  5.1\% &
  16 &
  0.4\% &
  218 &
  5.5\% \\
dialog &
  1600 &
  62 &
  3.9\% &
  26 &
  1.6\% &
  88 &
  5.5\%\\
          \hline
\end{tabular}
  \caption{\label{misaligntable}
    Summary of False Positives and False Negatives in LLM Decisions}
\end{table*}

The common elements of evaluation prompt are as follows:

\begin{center}
\line(1,0){218}
\end{center}

\noindent{\small
You are evaluating a response submitted for a specific function call task against a set of standards. Below is the data:

\noindent\text{[BEGIN DATA]}

\noindent\text{***}

\noindent\text{[Available Functions]}
\begin{verbatim}
{tools}
\end{verbatim}

\noindent\text{[Query]:}
\begin{verbatim}
{query}
\end{verbatim}

\noindent\text{[Ground Truth]:}
\begin{verbatim}
{ground_truth}
\end{verbatim}

\noindent\text{[Submission]:}
\begin{verbatim}
{response}
\end{verbatim}

\noindent\text{***}

\noindent\text{[Criterion]:}

\noindent\text{***}

\noindent\text{[END DATA]}

\noindent Does the submission meet the criteria? Begin by explaining your reasoning step by step in Korean, without immediately revealing the outcome. Subsequently, on a separate line, clearly indicate whether it is a "pass" or "fail". For clarity, repeat your final decision once more (without quotes or punctuation, literally).
}

\begin{center}
\line(1,0){218}
\end{center}

These format refers to \citet{zhou2023lima}. The placeholders \verb|{tools}|, \verb|{query}| and so on will be replaced by specific details from the actual case being evaluated. As in the data from \citet{zhou2023lima}, The "Query" and "Submission," which correspond to the input and output of the model being evaluated, are included in the prompt. Specifically, since the FunctionChat evaluation system targets function calling models, the "Available Functions" section has been added. Additionally, to enhance the alignment between the llm judge and the human who designed the evaluation, "Ground Truth" has been incorporated. The prompt for evaluating Tool Call type outputs additionally includes "Acceptable Arguments," which serve as supplementary indicators for "Ground Truth."

The "Criterion" has been established in four different categories according to our definition of the output types for the fc model. While \citet{} presented a 6-scale Likert score, we have provided criteria for a pass or fail status. Subsequently, the four criteria applied to each type of output will be detailed.

\bigskip

\bigskip

\bigskip

\begin{itemize}[leftmargin=*] 
    \item \textbf{Tool Call}
\end{itemize}

{
\small
\noindent [Criterion]: Accuracy in selecting the proper function, and generating the function name and argument values

\noindent Determine if the [Submission] is a "pass" or "fail". You are given a [Ground Truth] for each [Query], so you can refer to this for evaluating the response.

\noindent "pass"

\noindent - Selected the appropriate function and accurately named it.

\noindent - All keys in arguments match those presented in [Ground Truth].

\noindent - Each argument value matches the type specified in [Available Functions].

\noindent - Each value in arguments was created appropriately, as presented in [Ground Truth]. For string types, a [Submission] passes if its argument matches or refers to the same content or subject as [Ground Truth] or [Aacceptable arguments], even without an exact text match.

\noindent - If 'Only ground truth is allowed.' appears under [Aacceptable arguments], it means that only when the argument value exactly matches [Ground Truth] will it be considered a pass.

\noindent "fail" 

\noindent - Selection error: Did not select a function or selected a different function than the one presented in [Ground Truth].
\noindent - Function name error: Failed to accurately create the function name as presented in [Ground Truth] (different spelling).
\noindent - Argument key error: Created a key not presented, or different from those presented in [Ground Truth].

\noindent - Argument value type error: The type of the created argument value is inappropriate (not created as the type specified in the description of [Available Functions]).

\noindent - Logical error in argument value: The created argument value exceeds the permissible range as per [Ground Truth] and [Acceptable Arguments].

}

\begin{itemize}[leftmargin=*] 
    \item \textbf{Answer Completion}
\end{itemize}

{
\small
\noindent [Criterion]: Appropriateness of completing the answer without altering semantics, based on context

\noindent Determine if the [Submission] is a "pass" or "fail". In this submission, the role of the assistant is to convey the result returned by a specific function to the user. Instead of directly passing on data in JSON format, it should be paraphrased into conversational human utterance. It's important that the paraphrased content does not semantically differ from the tool's content. You are provided with a [Ground Truth] for each [Query], which you can use to evaluate the response. However, the [Ground Truth] is not the absolute and only answer. A slightly more concise response is also acceptable.

}

\begin{itemize}[leftmargin=*] 
    \item \textbf{Slot Question}
\end{itemize}

{
\small
\noindent [Criterion]: Occurrence of a proper slot filling question

\noindent Evaluate whether it's a "pass" or a "fail".

\noindent In the [Query], a user asks the assistant a question or makes a request, and in [Available Functions], there exists a suitable function to perform this task, but there is a lack of required parameter values needed to call the function. In this [Submission], the assistant is required to ask the user for any additional information necessary to invoke the appropriate function and complete the task. You are given a [Ground Truth] for each [Query], so you can refer to this for evaluating the response.

\noindent "pass"

\noindent - Appropriate questions for slot filling were asked. (It is not an issue if the 'function call' or 'tool call' item is null.)

\noindent "fail"

\noindent - Tool call missing required information.

\noindent - Tool call with incorrect information: Hallucinated values not found in the [Query].

\noindent - Fail of function selection: Called a different, inappropriate function instead of the one that should be called (as can be verified through the [Ground Truth]).
 
\noindent - Answered arbitrarily based on their own knowledge without considering the function call.

}

\begin{itemize}[leftmargin=*] 
    \item \textbf{Relevance Detection}
\end{itemize}

{
\small
\noindent [Criterion]: Detecting the relevance of the [Query] to the function call functionality or [Available Functions]

\noindent Evaluate whether it's a "pass" or a "fail".
 
\noindent In the [Query], it presents a scenario that does not necessitate a tool call. You are given a [Ground Truth] for each [Query], so you can refer to this for evaluating the response.

\noindent "pass"

\noindent - For user statements that didn't require a function call, the model leveraged its available knowledge to interact naturally.

\noindent - When there's a need for an external tool or real-time information beyond the capabilities of the language model, and the [Available Functions] don't cover these needs, it's explained that the feature isn't provided. Therefore, it's clarified that the question cannot be answered or the request cannot be fulfilled.

\noindent "fail"

\noindent - A tool was called improperly or unnecessarily.

\noindent - The task was not rejected despite requiring an external tool or real-time information not covered by the language model and [Available Functions], leading to an inaccurate claim that it could be performed or had been performed.

}

\section{Misalignment in Judgment between Human and LLM}
\label{misalignment}

Misalignment cases include instances where the LLM judged a submission as "pass" but it was actually a "fail" (False Positives, FP), and where the LLM judged as "fail" but it was actually a "pass" (False Negatives, FN). The quantity and proportion of items that were adjusted through qualitative review are as shown in Table~\ref{misaligntable}.

\end{document}